\documentclass[runningheads]{llncs}

\usepackage{eccv}

\usepackage{eccvabbrv}

\usepackage{graphicx}
\usepackage{float}
\usepackage{booktabs}
\usepackage{siunitx}
\usepackage{multirow}
\usepackage{adjustbox}
\usepackage{caption}

\usepackage[accsupp]{axessibility}  % Improves PDF readability for those with disabilities.

\usepackage{hyperref}

\usepackage{orcidlink}

\begin{document}

% \title{Agentic Shadow Removal via Physics-Oriented Candidate Selection}
% \title{Are Foundation Models the End of Low-Level Vision? A Shadow Removal Case}
% \title{Agentic Physics-Oriented Candidate Selection for Shadow Removal}
\title{Domain-Grounded Candidate Selection for Agentic Image Editing: A Shadow Removal Case}

\titlerunning{Agentic Shadow Removal}

\author{
Shilin Hu\inst{1}\orcidlink{0009-0006-6002-1403}\and
Jingyi Xu\inst{1}\orcidlink{0009-0003-5597-5534}\and
Dimitris Samaras\inst{1}\textsuperscript{$\dagger$}\orcidlink{0000-0002-1373-0294}\and
Hieu Le\inst{2}\textsuperscript{$\dagger$}\orcidlink{0000-0001-7855-2778}
}

\authorrunning{S.~Hu et al.}

\institute{
Stony Brook University, Stony Brook NY 11794, USA \and UNC Charlotte, Charlotte NC 28223, USA\\
\email{\{shilhu,jingyixu,samaras\}@cs.stonybrook.edu},
\email{hle40@charlotte.edu}
}

\maketitle
\begingroup
\renewcommand{\thefootnote}{}
\NoHyper
\footnotetext{\textsuperscript{$\dagger$} Equal advising.}
\endNoHyper
\endgroup

\begin{abstract}

Commercial vision-language models are reshaping computer vision, with visual priors broad enough to rival task-specific systems. This raises a natural question: do they reduce the need for classic, physics-informed low-level vision? We study this through shadow removal, a problem shaped by scene geometry, illumination, materials, and occluders, where paired shadow and shadow-free data are hard to collect at scale. We find that a commercial generative editor, used directly, can produce clean shadow-free edits that preserve surface texture and local appearance. However, this comes with a new failure mode: the same editor can regenerate scene content, hallucinate objects, or misread a shadow as material or geometry, producing plausible but physically wrong edits. We address this with an agentic candidate-selection pipeline: the editor generates a guided probe, an evaluator screens for major failures, retries when needed, samples multiple candidates, filters them, and selects a final result balancing shadow removal against scene preservation. Grounding this process in shadow-formation physics makes it more reliable: prompting the generator and evaluator to treat shadows as illumination effects caused by light occlusion, not material or object structure, measurably improves quality and consistency. On the ShadowRemovalRefine benchmark, our physics-oriented pipeline achieves a CDD of 0.0075, reducing CDD by at least 47\% over the strongest prior method. These results suggest that commercial vision-language models do not replace classic low-level vision priors; instead, such priors remain useful for constraining and steering physically underconstrained generation.

\end{abstract}

\section{Introduction}
\label{sec:intro}

\begin{figure}[!t]
    \centering
    \includegraphics[width=\linewidth]{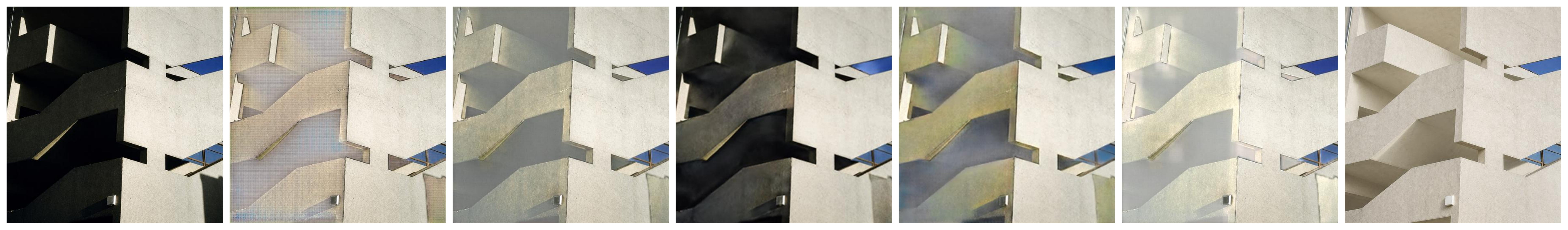}
    \par
    \includegraphics[width=\linewidth]{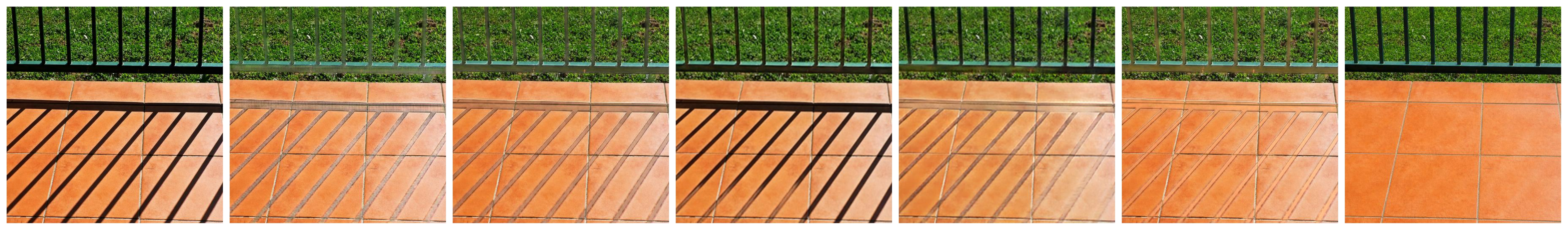}
    \par
    {\small
    \makebox[\linewidth][c]{%
        \makebox[0.142\linewidth][c]{Input}%
        \makebox[0.142\linewidth][c]{SID~\cite{le2019shadowdecomposition}}%
        \makebox[0.142\linewidth][c]{SF~\cite{guo2023shadowformer}}%
        \makebox[0.142\linewidth][c]{SD~\cite{guo2023shadowdiffusion}}%
        \makebox[0.142\linewidth][c]{I4S~\cite{li2023inpainting}}%
        \makebox[0.142\linewidth][c]{SRR~\cite{hu2025shadowrefine}}%
        \makebox[0.142\linewidth][c]{Ours}%
    }}
    \caption{
    \textbf{Complex real-world shadows.}
    Examples from the ShadowRemovalRefine benchmark~\cite{hu2025shadowrefine}.
    From left to right: input image, SID~\cite{le2019shadowdecomposition}, ShadowFormer~\cite{guo2023shadowformer}, ShadowDiffusion~\cite{guo2023shadowdiffusion}, Inpaint4Shadow~\cite{li2023inpainting}, SRR~\cite{hu2025shadowrefine}, and ours.
    Prior methods often leave residual shadows, retain visible shadow boundaries, or produce washed-out surface appearance, while our method achieves cleaner and more complete shadow removal.
    }
    \label{fig:teaser}
    \vspace{-1mm}
\end{figure}

Commercial vision-language and generative editing models~\cite{li2023blip2,liu2023visual,brooks2023instructpix2pix,zhang2023magicbrush,sheynin2024emuedit,zhao2024ultraedit} are reshaping computer vision. Trained on internet-scale data, they carry visual priors broad enough to edit photographs with a level of detail and flexibility that task-specific restoration models often struggle to match. This progress raises a pointed question for the low-level vision community: if a general-purpose editor can already produce clean, photorealistic edits, what role remains for classic, task-specific models?

We study this question through shadow removal, a natural test case: shadow appearance depends on scene geometry, lighting, materials, and occluders, and paired shadow and shadow-free training data are difficult to collect at scale. These challenges have motivated both physically inspired shadow formulations and paired-data restoration models~\cite{qu2017deshadownet,wang2018stacked,le2019shadowdecomposition,cun2020ghostfree,le2020shadowsegmentation,guo2023shadowformer,guo2023shadowdiffusion,li2023inpainting,hu2025shadowrefine}. We find that, given a mask and a text instruction, a commercial editor can produce clean shadow-free edits that often preserve surface texture and local appearance better than prior shadow-removal methods. But the same editor has no explicit notion of how a shadow forms, so a single query can fail in a specific way: it may remove a shadow by regenerating scene content, misread reduced illumination as material or geometry, or hallucinate new objects and textures into the target region, producing edits that look convincing while being physically wrong. We address this with an agentic candidate-selection system that detects and screens out these failures using simple shadow-formation physics: a shadow is an illumination effect caused by light occlusion rather than a material or object property.

Concretely, we formulate shadow removal as an agentic candidate-selection problem. Given an image and a shadow mask, our system generates a guided probe, evaluates it for major failures, conditionally retries, samples multiple candidate edits, filters unacceptable candidates, and selects a final output according to structured visual criteria. Sampling multiple candidates reduces dependence on any single stochastic generation and gives the evaluator diverse alternatives to compare, turning an unstable single-shot editor into a controllable shadow removal pipeline. This pipeline alone, however, only goes so far: an evaluator with no notion of shadow formation can still be misled by a candidate that looks clean but is physically incorrect. We therefore embed the same illumination-based shadow model in both generation and evaluation, prompting the editor and evaluator to treat shadows as blocked illumination rather than material or object structure. This shadow-formation grounding steers the pipeline toward candidates that remove reduced illumination while preserving scene layout, object boundaries, and material appearance, yielding a consistent gain over an otherwise identical pipeline without such grounding.

On the ShadowRemovalRefine (SRR) benchmark~\cite{hu2025shadowrefine}, our physics-oriented agentic pipeline reaches a CDD of 0.0075, reducing CDD by at least 47\% over the strongest prior method. Our ablations show that probing, failure detection, candidate filtering, and final selection improve reliability, and that shadow-formation grounding provides a consistent further gain over a generic agentic variant. Foundation models, in other words, do not retire classic low-level vision; they change its role, from modeling a phenomenon end-to-end to steering and correcting a much more powerful but physically underconstrained generator.
\section{Method}
\label{sec:method}

\begin{figure}[!t]
    \centering
    \includegraphics[width=0.8\linewidth]{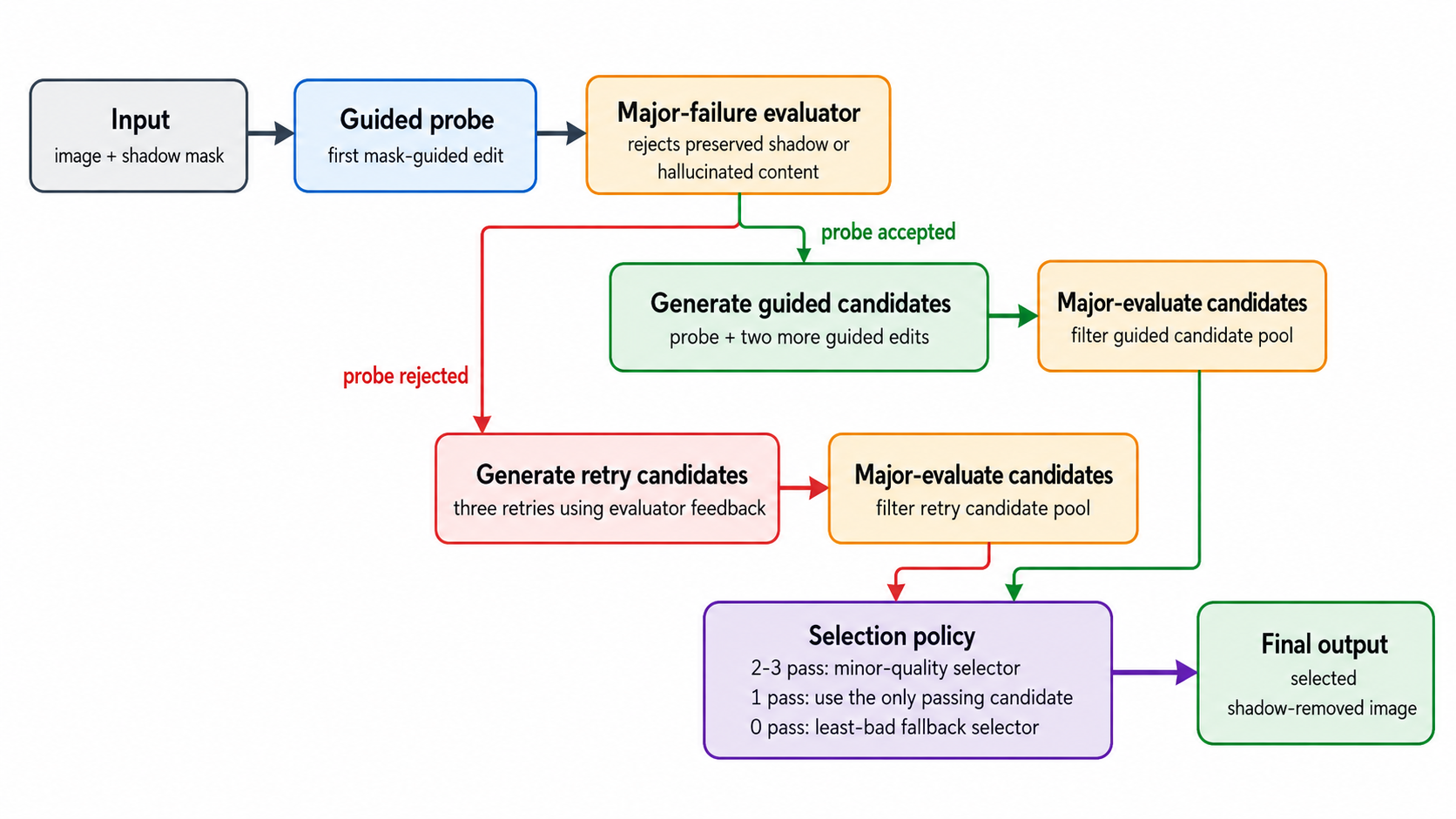}
    \caption{\textbf{Agentic candidate-selection pipeline.} The system first generates a guided probe and evaluates it for major failures. Accepted probes lead to additional guided candidates, while rejected probes trigger retry candidates using evaluator feedback. Candidate filtering and final selection produce the shadow-removed output.}
    \label{fig:pipeline}
\end{figure}

\begin{figure}[!t]
    \centering
    \includegraphics[width=0.8\linewidth]{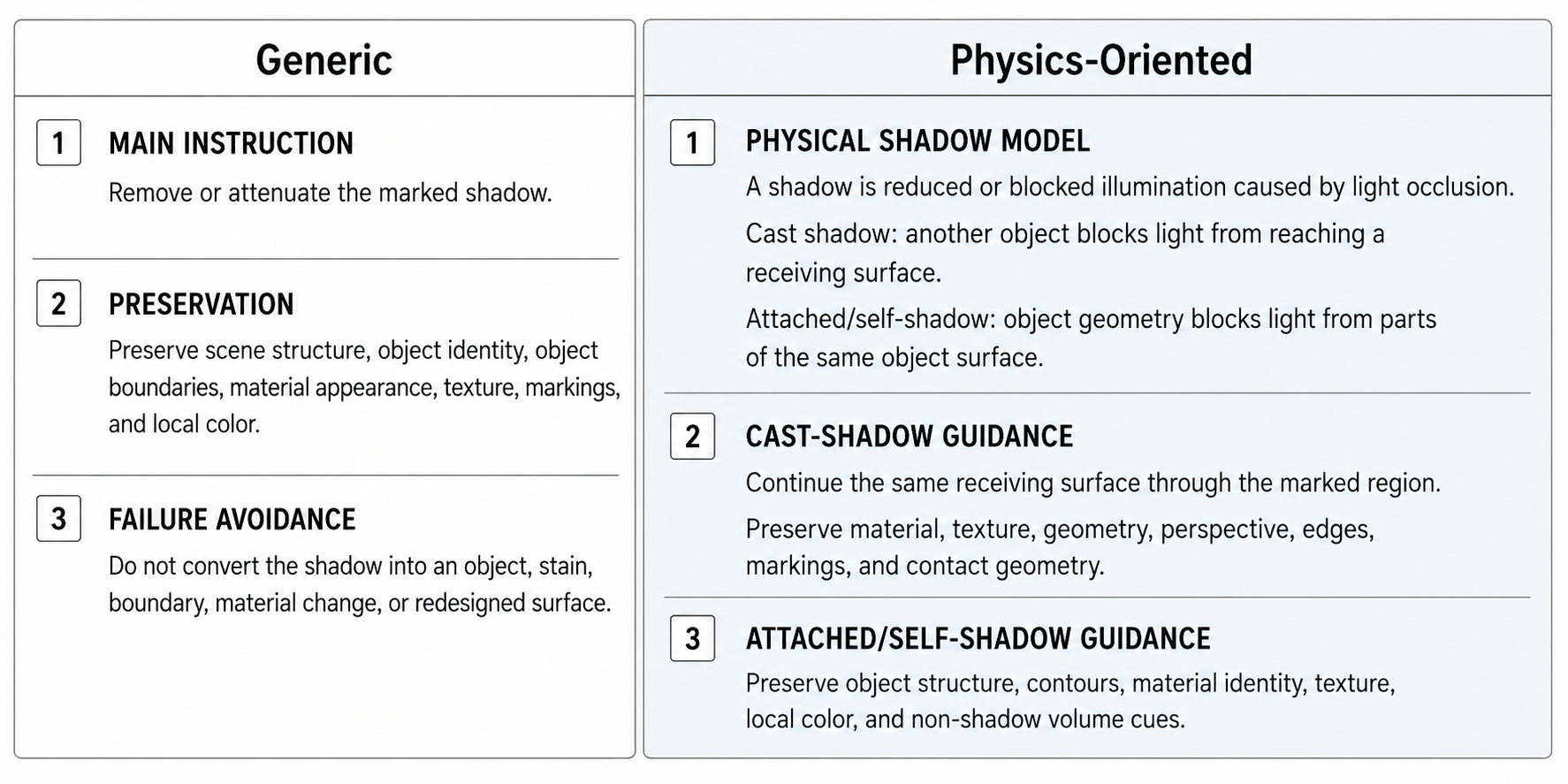}
    \caption{
    \textbf{Generic versus physics-oriented prompting.}
    Both settings use the same agentic candidate-selection pipeline.
    Physics-oriented prompting adds shadow-formation definitions, encouraging the generator and evaluator to treat shadows as illumination effects rather than material or object structure.
    }
    \label{fig:genericvsphysics}
\end{figure}

Given an input image $I$ and shadow mask $M$, our goal is to produce a shadow-removed image $\hat{I}$ while preserving scene layout, object structure, and non-shadow appearance. We use a generative image editor as a stochastic candidate generator and a multimodal evaluator as the agentic controller.

\paragraph{Guided probe and major-failure evaluation.}
As shown in \cref{fig:pipeline}, the pipeline begins with a guided probe: a single mask-guided edit generated from the input image, shadow mask, and shadow removal instruction. A major-failure evaluator rejects severe errors, including preserving the shadow as an object-like dark structure or hallucinating semantic content in the target region.

\paragraph{Candidate generation and selection.}
The probe outcome determines candidate generation. If the probe passes, we keep it and sample two additional guided candidates; otherwise, we generate three retry candidates using evaluator feedback. All candidates are filtered by the same major-failure evaluator. When multiple candidates pass, a minor-quality selector chooses the final output by balancing shadow removal with scene preservation; when one candidate passes, it is selected directly; when none pass, a fallback selector chooses the least-bad candidate.

\paragraph{Shadow-formation grounding.}
We evaluate two prompt settings under the same pipeline, as summarized in \cref{fig:genericvsphysics}. The generic setting uses direct shadow removal and preservation instructions. The physics-oriented setting adds simple shadow-formation grounding, prompting the generator and evaluator to treat shadows as illumination effects caused by light occlusion rather than material or object structure. This encourages removal of reduced illumination while preserving scene geometry and material appearance.

\section{Experiments and Results}
\label{sec:experiments}

\begin{table}[!t]
    \centering
    \caption{CDD comparison on the SRR benchmark~\cite{hu2025shadowrefine}. We report the mean, standard deviation, minimum, and maximum CDD across benchmark images, with all values multiplied by $10^3$ for display; lower values indicate better shadow removal. The best result in each column is \textbf{bolded}.}
    \label{tab:sota}
    \small
    \setlength{\tabcolsep}{8pt}
    \renewcommand{\arraystretch}{1.12}
    \begin{tabular}{
        l
        S[table-format=2.2]
        S[table-format=3.2]
        S[table-format=1.2]
        S[table-format=3.2]
    }
        \toprule
        Method & {Mean} & {Std} & {Min} & {Max} \\
        \midrule
        SID~\cite{le2019shadowdecomposition}
            & 38.00 & 45.05 & 1.66 & 353.64 \\
        ShadowFormer~\cite{guo2023shadowformer}
            & 33.63 & 52.74 & 0.41 & 427.34 \\
        ShadowDiffusion~\cite{guo2023shadowdiffusion}
            & 89.72 & 130.54 & 0.29 & 650.26 \\
        Inpaint4Shadow~\cite{li2023inpainting}
            & 31.57 & 48.39 & 0.43 & 402.44 \\
        SRR~\cite{hu2025shadowrefine}
            & 14.06 & 26.08 & 0.24 & 215.40 \\
        \midrule
        Ours-Generic
            & 10.33 & 28.72 & \bfseries 0.01 & 298.38 \\
        Ours-Physics
            & \bfseries 7.45 & \bfseries 20.65 & \bfseries 0.01 & \bfseries 200.19 \\
        \bottomrule
    \end{tabular}
\end{table}

\paragraph{Implementation details.}
We use GPT-image-2~\cite{openai2026imagesvision,openai2026models} for candidate generation and GPT-5-mini from the GPT-5 family~\cite{openai2025gpt5systemcard} for multimodal evaluation and selection, with all calls made through the OpenAI API. Each full benchmark run processes the SRR benchmark once, takes roughly 14 hours, and uses up to four image-editing calls per case plus evaluator/selector calls.

\paragraph{Evaluation protocol.}
We evaluate on the ShadowRemovalRefine (SRR) benchmark~\cite{hu2025shadowrefine}. We first normalize all benchmark images and masks to $512 \times 512$, and all methods are evaluated on this normalized benchmark set. The benchmark contains 400 images; one image was rejected by the API, leaving 399 evaluated cases. Following SRR, we report Color Distribution Difference (CDD), which measures color-distribution discrepancy across annotated shadow boundaries; lower CDD indicates better shadow removal and boundary consistency. We compare against SID~\cite{le2019shadowdecomposition}, ShadowFormer~\cite{guo2023shadowformer}, ShadowDiffusion~\cite{guo2023shadowdiffusion}, Inpaint4Shadow~\cite{li2023inpainting}, and SRR~\cite{hu2025shadowrefine}. Because image generation is stochastic, we run each agentic variant twice and report averaged statistics across the two full runs.

\begin{figure}[!t]
    \centering
    \includegraphics[width=\linewidth]{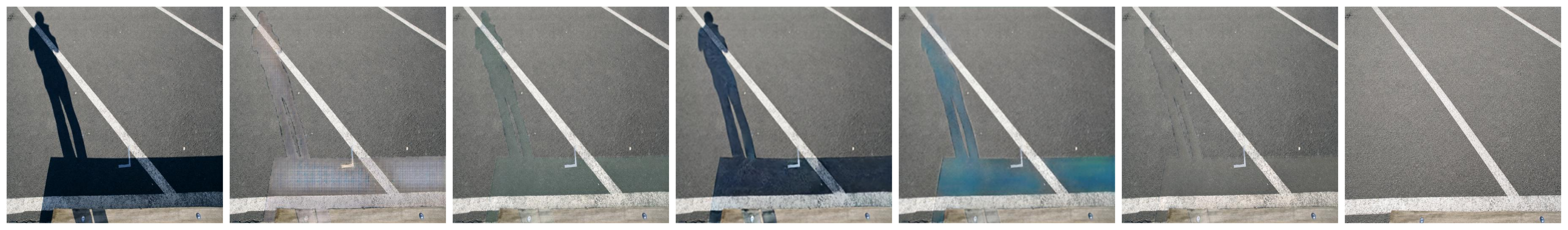}
    \par
    \includegraphics[width=\linewidth]{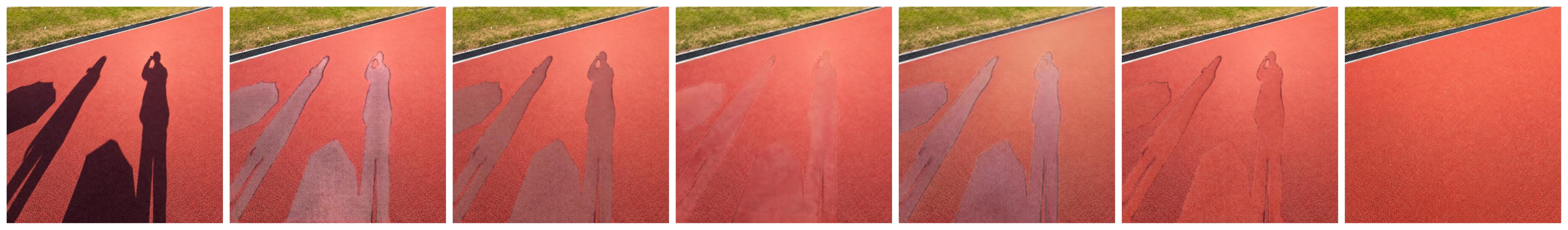}
    \par
    \includegraphics[width=\linewidth]{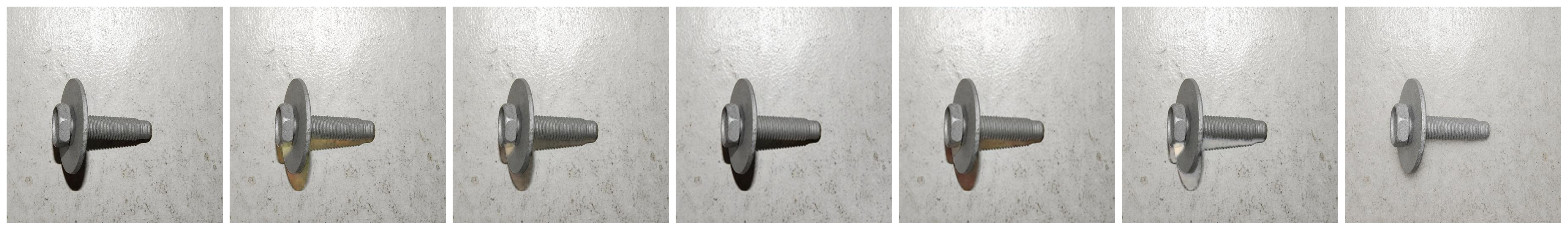}
    \par
    \includegraphics[width=\linewidth]{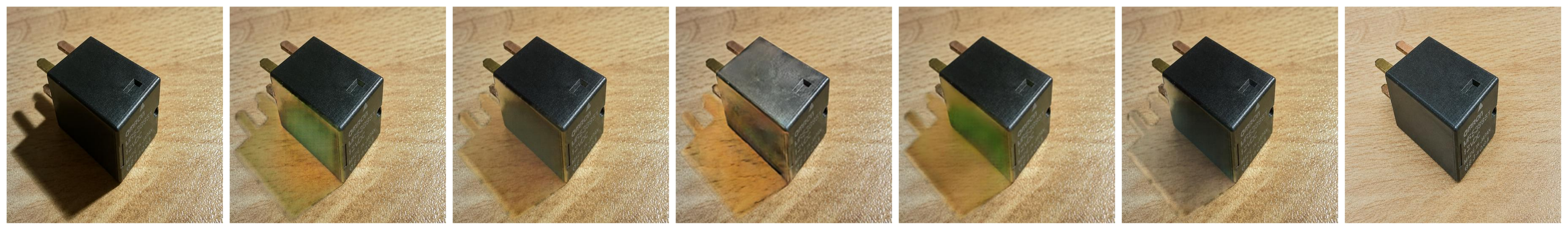}
    \par
    \includegraphics[width=\linewidth]{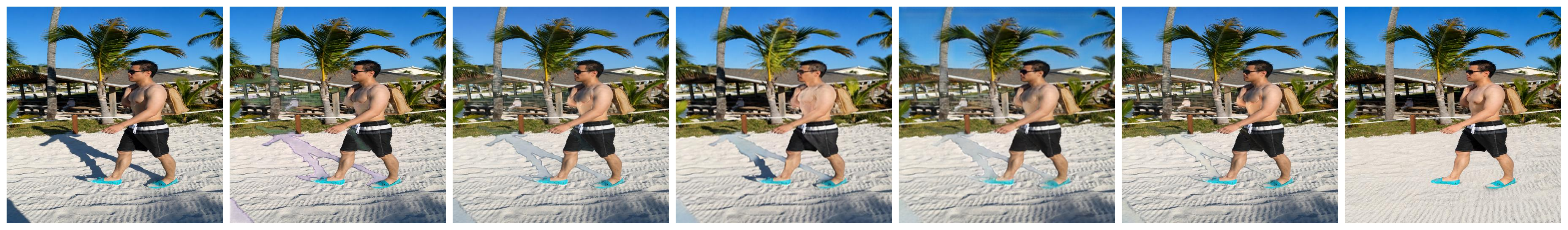}
    \par
    \includegraphics[width=\linewidth]{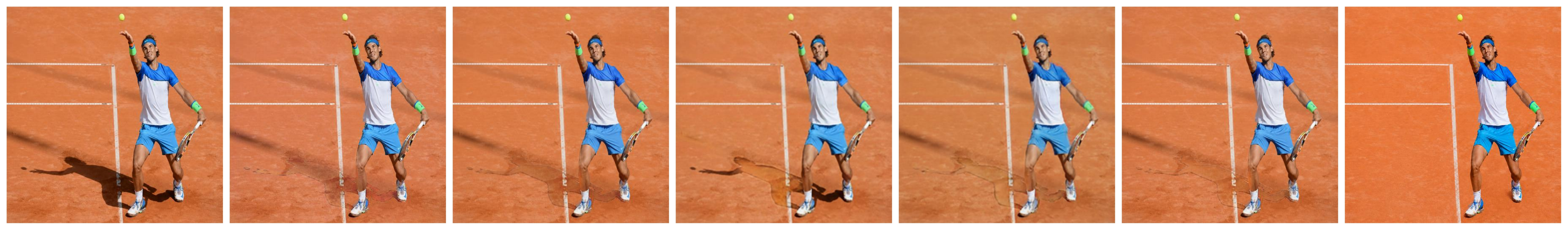}
    \par
    \includegraphics[width=\linewidth]{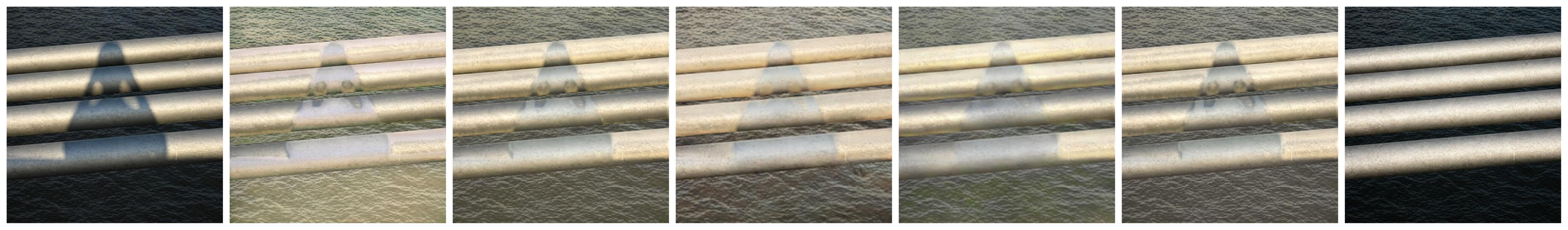}
    \par
    \includegraphics[width=\linewidth]{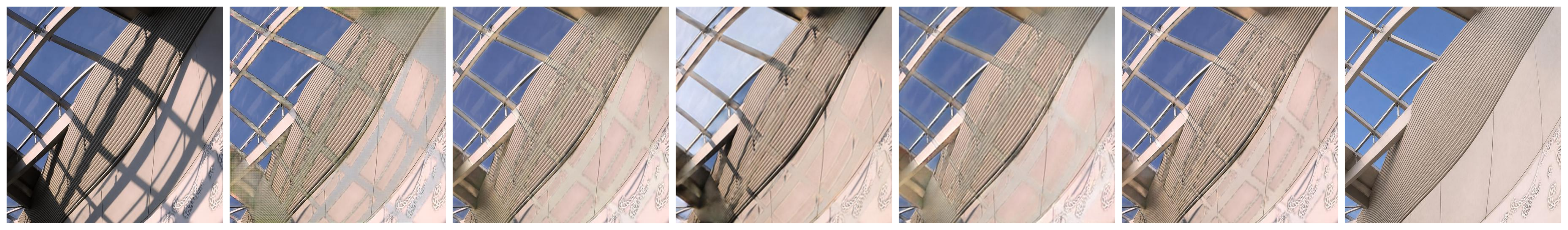}
    \par
    {\small
    \makebox[\linewidth][c]{%
        \makebox[0.142\linewidth][c]{Input}%
        \makebox[0.142\linewidth][c]{SID~\cite{le2019shadowdecomposition}}%
        \makebox[0.142\linewidth][c]{SF~\cite{guo2023shadowformer}}%
        \makebox[0.142\linewidth][c]{SD~\cite{guo2023shadowdiffusion}}%
        \makebox[0.142\linewidth][c]{I4S~\cite{li2023inpainting}}%
        \makebox[0.142\linewidth][c]{SRR~\cite{hu2025shadowrefine}}%
        \makebox[0.142\linewidth][c]{Ours}%
    }}
    \caption{
    \textbf{Qualitative comparison.}
    From left to right: input image, SID~\cite{le2019shadowdecomposition}, ShadowFormer~\cite{guo2023shadowformer}, ShadowDiffusion~\cite{guo2023shadowdiffusion}, Inpaint4Shadow~\cite{li2023inpainting}, SRR~\cite{hu2025shadowrefine}, and ours.
    Across diverse shadows and surface materials, our method more consistently reduces residual shadow regions and boundary artifacts while preserving plausible scene appearance.
    }
    \label{fig:qual}
    \vspace{-1mm}
\end{figure}

\subsection{Quantitative comparison}
As shown in \cref{tab:sota}, Ours-Physics achieves the lowest mean CDD among all evaluated methods, reaching 0.0075 and reducing CDD by at least 47.0\% compared with the strongest prior method, SRR. Ours-Generic also improves over prior methods, indicating that combining a generative image editor with candidate selection is effective, while shadow-formation grounding further improves both mean CDD and standard deviation.

\subsection{Qualitative comparison}
\Cref{fig:qual} compares our method with prior shadow removal approaches on SRR examples with varied shadow strengths, surfaces, and boundary structures. Existing methods often attenuate the shadow but leave visible boundary traces, residual dark regions, or washed-out surface appearance. Our method more consistently removes the dominant shadow region and suppresses shadow-edge artifacts, producing outputs that better match the surrounding scene appearance.

\subsection{Ablation analysis}
\label{sec:ablation}

\begin{table}[!t]
    \centering
    \caption{Ablation of the agentic candidate-selection pipeline on the SRR benchmark~\cite{hu2025shadowrefine}. We report CDD multiplied by $10^3$ for display; lower values indicate better shadow removal. The best mean and standard deviation are \textbf{bolded}.}
    \label{tab:ablation}
    \small
    \setlength{\tabcolsep}{8pt}
    \renewcommand{\arraystretch}{1.12}
    \begin{tabular}{
        ll
        S[table-format=2.2]
        S[table-format=2.2]
        S[table-format=1.2]
        S[table-format=3.2]
    }
        \toprule
        Prompt & Variant & {Mean} & {Std} & {Min} & {Max} \\
        \midrule
        \multirow{4}{*}{Generic}
            & Probe
            & 12.07 & 35.08 & 0.02 & 362.31 \\
            & Major-retry
            & 11.77 & 32.90 & 0.02 & 348.03 \\
            & Selector-worst
            & 11.39 & 33.73 & 0.02 & 442.28 \\
            & Selected
            & 10.33 & 28.72 & 0.01 & 298.38 \\
        \midrule
        \multirow{4}{*}{Physics}
            & Probe
            & 9.19 & 27.72 & 0.02 & 385.78 \\
            & Major-retry
            & 8.16 & 22.85 & 0.02 & 192.43 \\
            & Selector-worst
            & 8.75 & 25.45 & 0.01 & 335.35 \\
            & Selected
            & \bfseries 7.45 & \bfseries 20.65 & 0.01 & 200.19 \\
        \bottomrule
    \end{tabular}
\end{table}

\begin{figure}[!t]
    \centering
    \includegraphics[width=0.5\linewidth]{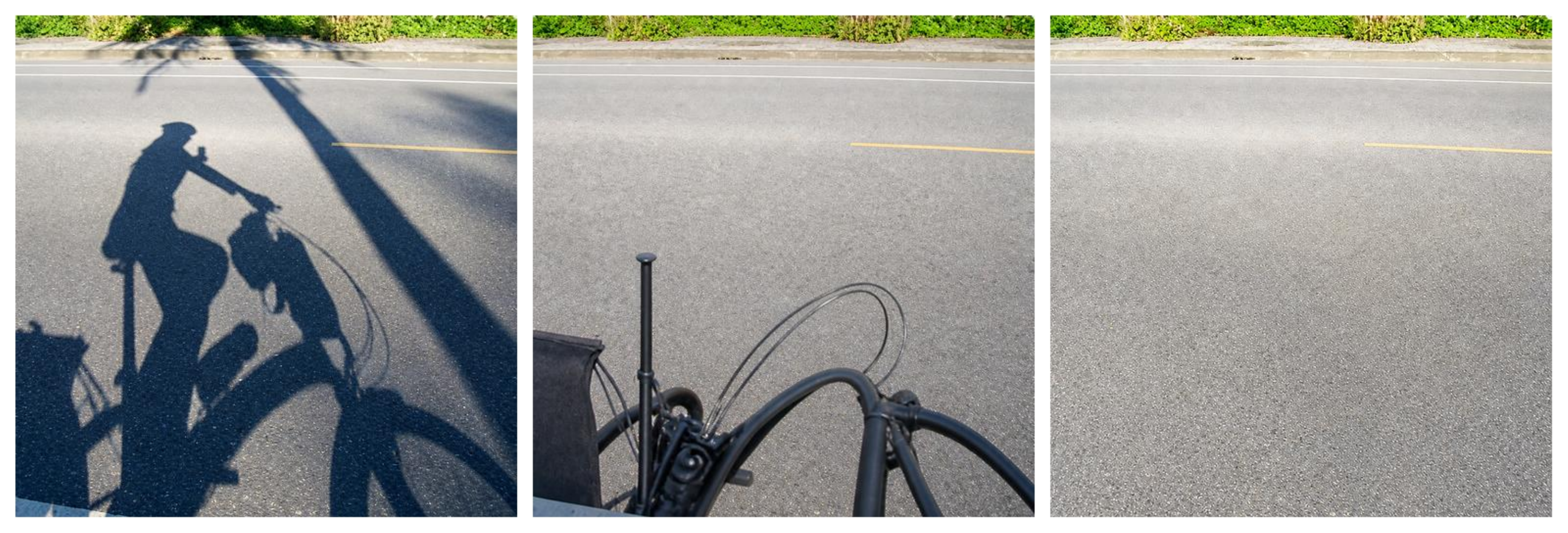}%
    \includegraphics[width=0.5\linewidth]{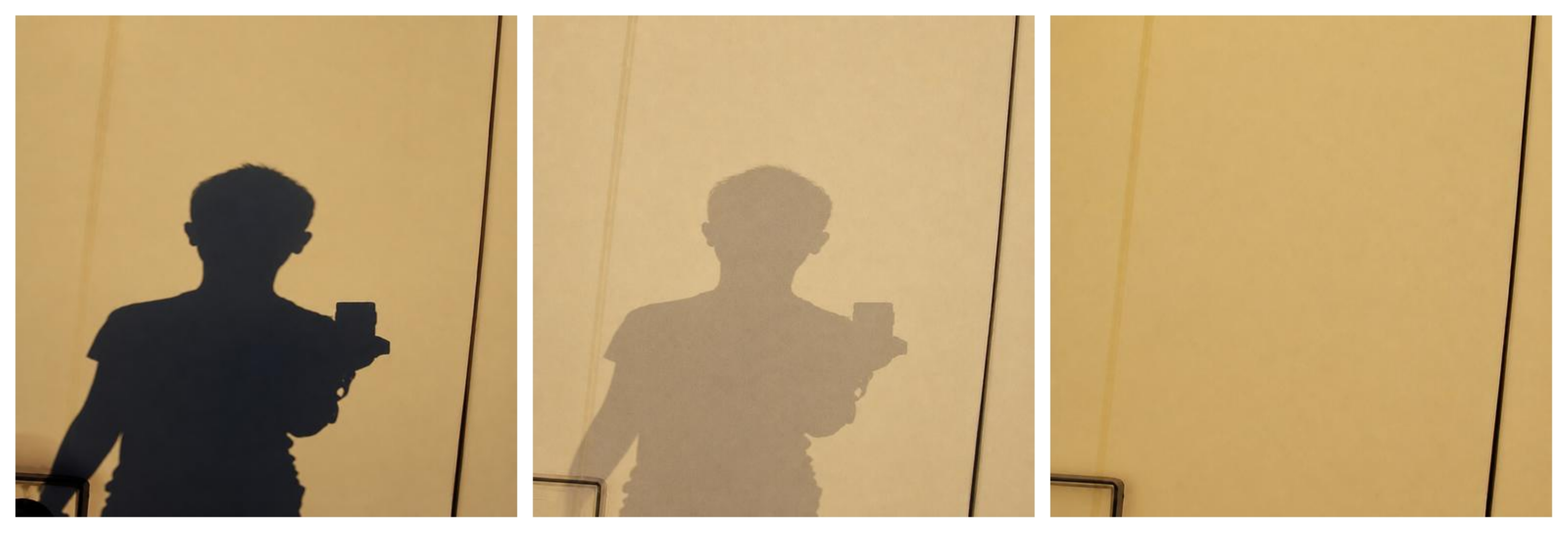}
    \par
    {\small
    \makebox[\linewidth][c]{%
        \makebox[0.166\linewidth][c]{Input}%
        \makebox[0.166\linewidth][c]{Probe}%
        \makebox[0.166\linewidth][c]{Retry}%
        \makebox[0.166\linewidth][c]{Input}%
        \makebox[0.166\linewidth][c]{Probe}%
        \makebox[0.166\linewidth][c]{Retry}%
    }}
    \caption{
    \textbf{Effect of major-failure retry.}
    The major-failure evaluator targets severe probe errors such as residual shadow structure and object-like hallucination from the shadow region.
    In these examples, rejected probes trigger retry generation, producing cleaner candidates before final candidate selection.
    }
    \label{fig:major}
\end{figure}
\begin{figure}[!t]
    \centering
    \includegraphics[width=\linewidth]{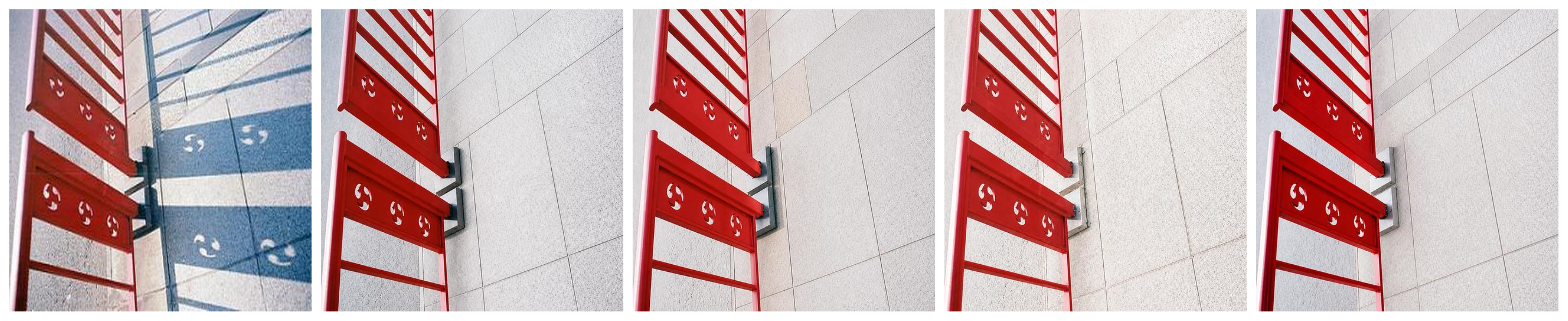}
    \par
    \includegraphics[width=\linewidth]{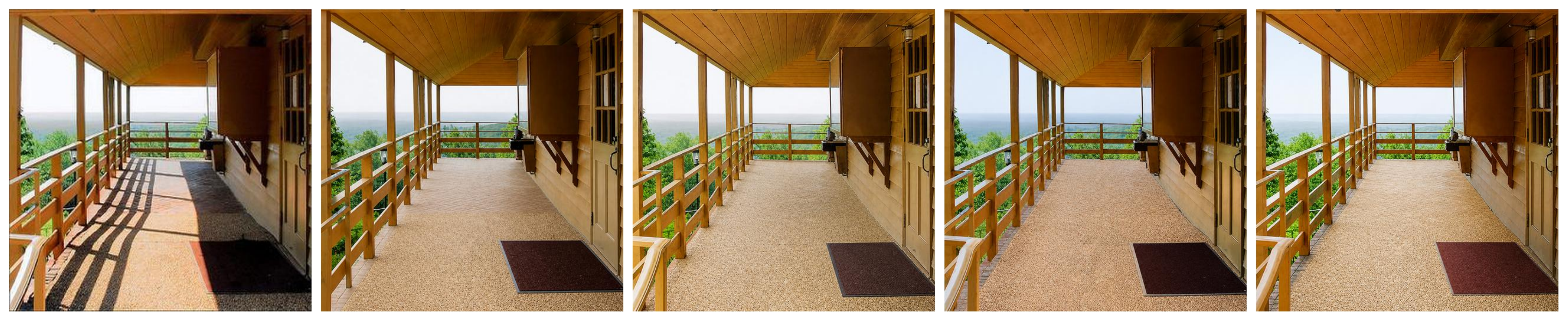}
    \par
    \includegraphics[width=\linewidth]{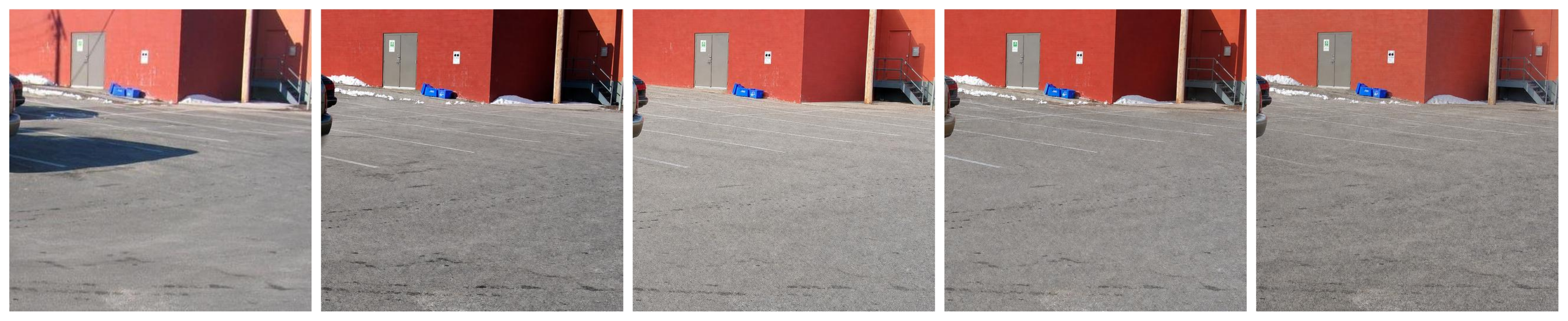}
    \par
    {\small
    \makebox[\linewidth][c]{%
        \makebox[0.20\linewidth][c]{Input}%
        \makebox[0.20\linewidth][c]{Generic}%
        \makebox[0.20\linewidth][c]{Physics Cand.}%
        \makebox[0.20\linewidth][c]{Physics Cand.}%
        \makebox[0.20\linewidth][c]{Physics selected}%
    }}
    \caption{
    \textbf{Effect of shadow-formation grounding and candidate selection.}
    The generic pipeline can make task-level mistakes, while physics-oriented prompting produces stronger candidate edits and final selection chooses the cleaner output.
    }
    \label{fig:genvsphycand}
    % \vspace{-1mm}
\end{figure}

\Cref{tab:ablation} evaluates the main stages of the candidate-selection pipeline. In both prompt settings, the selected output improves over the initial probe, showing that major-failure filtering, retry, and final selection are more reliable than using a single stochastic edit. The physics-oriented setting further improves over the generic setting, suggesting that shadow-formation grounding helps the generator and evaluator prefer candidates with cleaner shadow removal.

\Cref{fig:major} shows representative retry cases where the major-failure evaluator rejects a failed probe and triggers a cleaner retry candidate. \Cref{fig:genvsphycand} further illustrates the combined effect of shadow-formation grounding and candidate selection: the generic pipeline can still leave residual shadows, introduce boundary artifacts, or make task-level mistakes when interpreting dark regions, while physics-oriented prompting produces a stronger candidate set and final selection chooses the candidate that best balances shadow removal with scene consistency.
\section{Discussion}
\label{sec:discussion}

\begin{figure}[!t]
  \centering

  \begin{adjustbox}{valign=t,minipage=0.46\linewidth}
    \centering
    \captionof{table}{Non-shadow preservation on the SRR benchmark~\cite{hu2025shadowrefine}. RGB RMSE is computed outside the target shadow mask, measuring whether unmasked regions remain unchanged; lower is better.}
    \label{tab:nonshadow_rmse}

    \small
    \setlength{\tabcolsep}{3pt}
    \renewcommand{\arraystretch}{1.08}
    \begin{tabular}{
      @{}l
      S[table-format=2.2]
      S[table-format=2.2]
      S[table-format=2.2]@{}
    }
      \toprule
      Method & {Mean} & {Std} & {Median} \\
      \midrule
      SID~\cite{le2019shadowdecomposition}
          & \bfseries 3.44 & \bfseries 1.94 & \bfseries 3.08 \\
      SF~\cite{guo2023shadowformer}
          & 8.28 & 3.09 & 7.93 \\
      SD~\cite{guo2023shadowdiffusion}
          & 18.39 & 8.31 & 17.10 \\
      I4S~\cite{li2023inpainting}
          & 16.38 & 6.39 & 15.15 \\
      SRR~\cite{hu2025shadowrefine}
          & 7.55 & 2.85 & 7.12 \\
      \midrule
      Ours-Generic
          & 25.75 & 11.80 & 23.86 \\
      Ours-Physics
          & 28.79 & 11.44 & 27.06 \\
      \bottomrule
    \end{tabular}
  \end{adjustbox}
  \hfill
  \begin{adjustbox}{valign=t,minipage=0.52\linewidth}
    \centering
    \includegraphics[width=\linewidth]{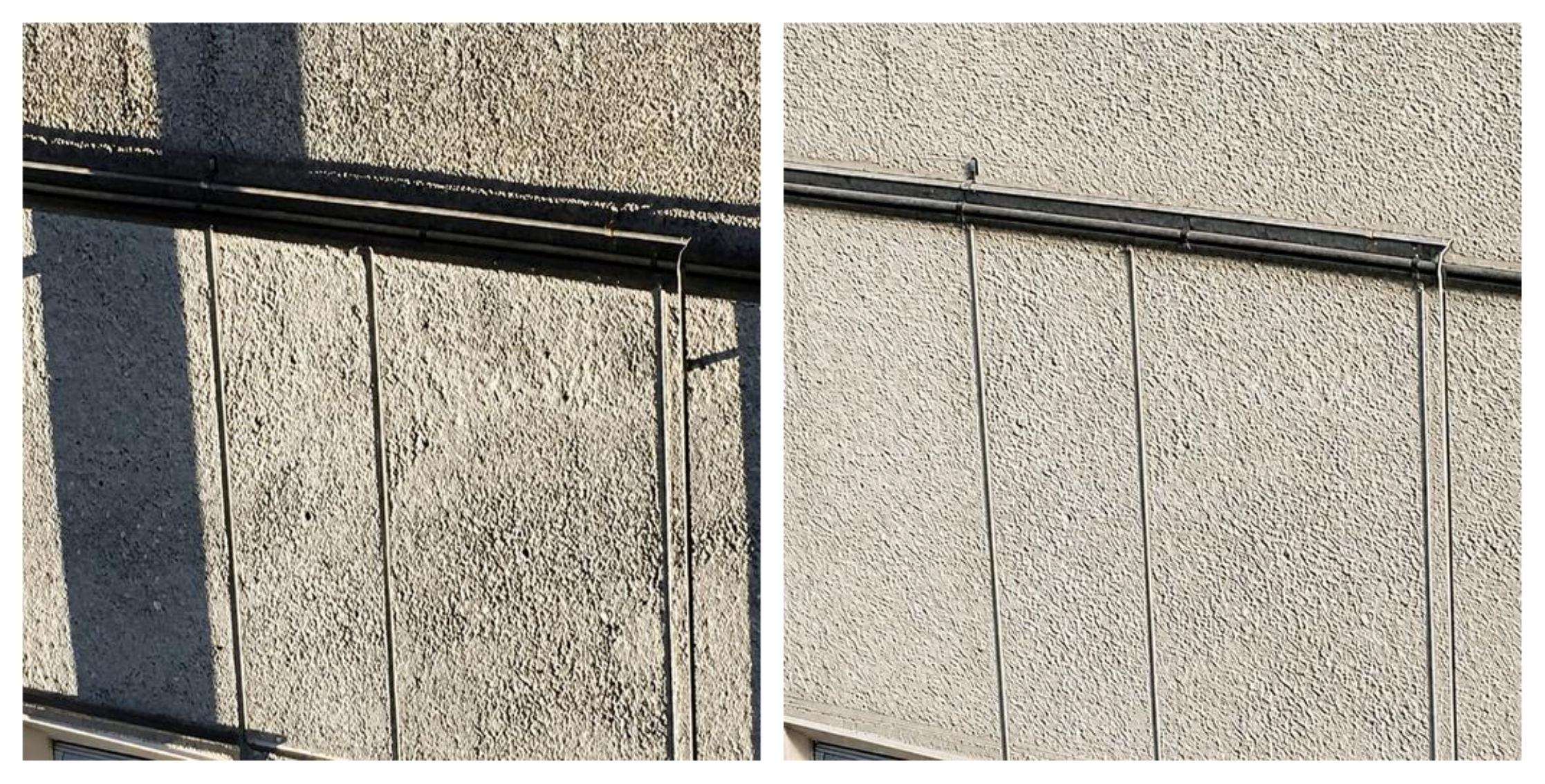}
    \par
    {\small
      \makebox[0.49\linewidth]{Input}%
      \makebox[0.49\linewidth]{Selected}%
    }

    \captionof{figure}{\textbf{Limitation.}
    Our method has two main failure modes: dark-region geometry change and overall appearance shift. This example shows both: the dark gap geometry is modified, and the unmasked wall is smoothed with a shifted color tone.}
    \label{fig:nonshadow_limitation}
  \end{adjustbox}
% \vspace{-2mm}
\end{figure}

\paragraph{Limitations.}
Although our pipeline improves shadow-boundary CDD, it does not always preserve the entire non-shadow scene. As shown in \cref{tab:nonshadow_rmse,fig:nonshadow_limitation}, generative editing can introduce appearance shifts outside the target shadow mask, including changes in color tone, surface smoothness, and fine structure. It can also modify dark geometric regions when they are visually entangled with the shadow. These failures suggest that candidate selection improves reliability but does not fully solve preservation when using general-purpose generative editors.

\paragraph{Conclusion and future work.}
We present a domain-grounded candidate-selection pipeline for controllable shadow removal with general-purpose generative image editors. By combining probing, major-failure retry, candidate filtering, final selection, and shadow-formation grounding, the method achieves state-of-the-art CDD on the SRR benchmark without training a task-specific restoration model. 
Future work will refine selected outputs with additional learned correction or fine-tuning stages to improve preservation and reliability. We also plan to curate a large-scale real-world shadow removal benchmark with paired shadow and shadow-free examples, covering more complex scenes, materials, illumination conditions, and shadow configurations than existing datasets.

% \section*{Acknowledgements}
% Please insert your acknowledgments here.

\bibliographystyle{splncs04}
\bibliography{main}
\end{document}